\documentclass[a4paper,conference]{IEEEtran}

\usepackage[pdftex]{graphicx}
\usepackage{verbatim}
\usepackage[graphicx]{realboxes}
\usepackage{adjustbox}
\usepackage{tablefootnote}
\usepackage{tabularx}
\usepackage{multirow, makecell}
\usepackage{array}
\usepackage{amsfonts}
\usepackage{diagbox}
\usepackage{amssymb}
\usepackage{enumerate}
\usepackage{enumitem}
\usepackage{cite}
\usepackage{amsmath}
\usepackage{amsmath}
\usepackage[caption=false,font=footnotesize]{subfig}
\usepackage{url}

\begin{document}

\title{Factorization Approach for Sparse Spatio-Temporal Brain-Computer Interface}

\author{\IEEEauthorblockN{Byeong-Hoo Lee$^1$, Jeong-Hyun Cho$^1$, Byoung-Hee Kwon$^1$, and Seong-Whan Lee$^2$ }
\IEEEauthorblockA{$^1$Department of Brain and Cognitive Engineering, Korea University, Seoul, South Korea \\$^2$Department of Artificial Intelligence, Korea University, Seoul, South Korea\\
bh\_lee@korea.ac.kr, jh\_cho@korea.ac.kr, bh\_kwon@korea.ac.kr, sw.lee@korea.ac.kr}
}
% make the title area
\maketitle

\begin{abstract}
Recently, advanced technologies have unlimited potential in solving various problems with a large amount of data. However, these technologies have yet to show competitive performance in brain-computer interfaces (BCIs) which deal with brain signals. Basically, brain signals are difficult to collect in large quantities, in particular, the amount of information would be sparse in spontaneous BCIs. In addition, we conjecture that high spatial and temporal similarities between tasks increase the prediction difficulty. We define this problem as sparse condition. To solve this, a factorization approach is introduced to allow the model to obtain distinct representations from latent space. To this end, we propose two feature extractors: A class-common module is trained through adversarial learning acting as a generator; Class-specific module utilizes loss function generated from classification so that features are extracted with traditional methods. To minimize the latent space shared by the class-common and class-specific features, the model is trained under orthogonal constraint. As a result, EEG signals are factorized into two separate latent spaces. Evaluations were conducted on a single-arm motor imagery dataset. From the results, we demonstrated that factorizing the EEG signal allows the model to extract rich and decisive features under sparse condition.
\end{abstract}
\IEEEpeerreviewmaketitle

\section{Introduction}
The human brain has an incredible problem-solving capability and infinite potential. Inspired by the brain, deep neural networks have shown outstanding performance in pattern recognition tasks such as image processing \cite{chen2021points, lee1997new, kao2021activity, lee2022human}, speech processing \cite{lee2021voicemixer, liu2021audio, lee2022duration, liu2021mutual, Lee_Yoon_Noh_Kim_Lee_2021}, and language processing \cite{devlin2018bert, lee2021korealbert, yang2021pose}. Recently, they have shown remarkable performance in detecting human intentions from brain signals \cite{lee2022quantifying, cho2021neurograsp}. Particularly, brain-computer interface (BCI) utilizes deep neural networks to develop a communication pathway between brain and external devices using brain signals \cite{won2017motion, wheelchair, chen2016high, suk2014predicting, lee2022motor}. BCI collects brain signals in invasive and non-invasive ways; In invasive BCI, brain signals are obtained from electrodes implanted directly into the brain and thus have relatively high quality, but it requires brain surgery \cite{hill2012recording, lee2015subject}. Non-invasive BCI mainly uses electrodes placed on the scalp to collect brain signals which are called an electroencephalogram (EEG). EEG signals are commonly used brain signals because those signals can be obtained without surgical approach. EEG signals have poor spatial resolution and low signal-to-noise ratio that are the main obstacles of non-invasive BCI.

Particularly, in the case of spontaneous BCI in which the user voluntarily generates control signals, the obstacles mentioned above are prominent. Therefore, the amplitude of the signals is low and the information is a form of a harmonic neural population firing which is not explicit. Many studies have developed paradigms to induce less noisy and high-quality EEG signals \cite{mcavinue2008measuring, sousa2017pure, DaSalla, won2017motion, zhang2017hybrid,lee2020classification}. Spontaneous BCI induces the user to produce valid control signals following paradigms; motor imagery (MI), visual imagery (VI), and speech imagery (SI). MI is a dynamic state in which the movements are rehearsed internally in the mind without actual movements \cite{bci3, decety1996neurophysiological, jeong2020decoding}. Thus, participants are asked to imagine specific muscle movements according to tasks. VI utilizes EEG signals that are generated during visual imagination. Participants consistently imagine specific images to generate control signals \cite{kwon2020decoding}. SI refers to speaking in mind without actual speaking \cite{dasalla2009single}. Several datasets were collected and publicly opened for decoding intention in EEG signals relied on these paradigms \cite{Nguyen, jeong2020multimodal, competition, kwon2020decoding, lee2021decoding}. Over the decades, numerous studies have been developed based on machine learning \cite{eegnet, deepconvnet, deeplearning, OYKwon}. They focused on extracting spatial and temporal features to obtain implicit representations of EEG signals. Especially, since EEG signals have high temporal resolution, recent studies have focused on extracting plenty of temporal features \cite{deeplearning2, JHJeong, MCNN}. Therefore, in the case of datasets with distinct regional differences, they achieved remarkable performance improvement. On the other hand, strategies for extracting spatial feature are necessary when the dataset uses only small regions of the brain such as MI tasks in a single-arm \cite{jeong2020multimodal} and SI \cite{Nguyen}. Therefore, imagery tasks within a single-arm would be difficult to distinguish using existing methods hence, EEG signals contain sparse spatio-temporal features.

\begin{figure*}[!t]
  \centerline{\includegraphics[scale = 1.2]{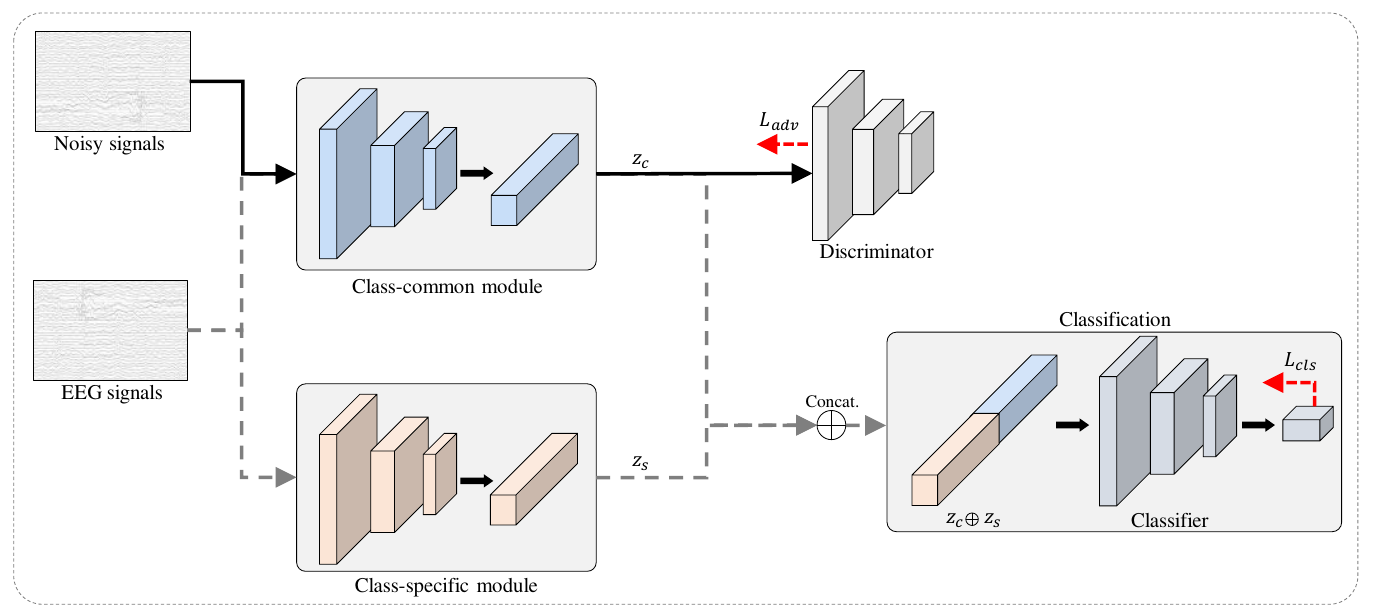}}
  \caption{The overall framework of the proposed method. Two modules were designed for extracting class-common and class-specific features, respectively. In particular, adversarial learning was applied to extract class-common features $z_c$. To this end, noise signals were utilized as an input of the module to extract fake features. Class-specific features $z_s$ are extracted using classification loss. The two types of features are concatenated, and the classifier receives the features as an input. Thus the classifier considers the two separate latent spaces and extract distinct features from them under sparse condition.}
\end{figure*}

In this paper, a factorization approach is proposed to acquire implicit representations of EEG signals. We conjecture that EEG signals generated in small regions of the brain have sparse information which is defined as sparse condition in this paper. Therefore strategic feature extraction should be applied because of little room for spatio-temporal features. Our approach is to explicitly factorize EEG signals into common and specific features to obtain discriminative representations against datasets under sparse condition. As a result, we designed two modules to extract class-common and class-specific features respectively. Class-common module learns common features of EEG signals through adversarial learning. Unlike other studies, we did not include a generator since the goal is to extract explicitly different types of features and hence no explicit/implicit modeling of the underlying input data distribution was required. Features from the modules are concatenated and fed into classifier for prediction. We conducted ablation studies to confirm the effectiveness of each design choice.

In summary, the main contributions of this paper are as follows: 1) We demonstrated that factorization is efficient for classifying EEG signals under sparse condition. To the best of our knowledge, it is the first attempt to explicitly factorize EEG signals for decoding user intentions. 2) An adversarial learning regime without the generator was introduced to obtain common features of EEG signals. Through this, the model can obtain separate latent spaces that enable the classifier to consider distinct representations of EEG signals. 3) We demonstrated that class-common and class-specific features are individually meaningless, but jointly use of the features improves classification performance.

\section{Related Works}

\subsection{MI Classification}
Several studies have contributed to tackling unsatisfactory classification performance. To obtain representations of EEG signals, Lu \textit{et al.} \cite{Lu} proposed a restricted Boltzmann machine-based network considering non-stationary properties of EEG signals. Ang \textit{et al.} \cite{FBCSP} developed filter bank to consider representations from different frequency range. The pipeline of this study has inspired other deep learning studies. Sakhavi \textit{et al.} \cite{sakhavi} presented a CNN architecture to extract diverse temporal representations based on \cite{FBCSP} exploiting high temporal resolution of EEG signals \cite{burle2015spatial}. Schirrmeister \textit{et al.} \cite{deepconvnet} proposed different depth of CNNs to explore multi-view classification. Furthermore, they described how the convolution works on the EEG signals by providing visualization. One of their contributions is revealed that band power features are efficient for MI classification. With the development of CNN-based networks, a study to control the number of parameters was also conducted by Lawhern \textit{et al.} \cite{eegnet}. They adopted depth-wise convolution and separable convolution to prove that a small number of parameters can achieve similar performance as existing methods. Amin \textit{et al.} \cite{MCNN} designed multiple CNN architectures for multi-view classification using MI dataset. Different depths allow the classifier to consider multi-level features. 

\subsection{SI Classification}
DaSalla \textit{et al.} \cite{DaSalla} introduced common spatial pattern (CSP) to obtain spatial representations for single-trial SI classification. Channel selection is efficient in extracting spatial features as demonstrated by Torres-Garcia \textit{et al.} \cite{Torres-Garcia}. Ngyuyen \textit{et al.} \cite{Nguyen} implemented Riemannian manifold with support vector machine \cite{lee2003pattern} for SI classification. 

However, these studies used datasets that involve relatively distinct brain regions which are an advantage for classification. In this study, we used a single-arm MI dataset \cite{jeong2020multimodal} that contains single-arm movement imagery tasks. Therefore, classes share only narrow brain regions that are an obstacle to improving classification performance. To the best of our knowledge, no one attempts to solve this problem yet, and the proposed method achieved performance improvement in \cite{jeong2020multimodal}.

\begin{figure}[!t]
  \centerline{\includegraphics[scale = 1.3]{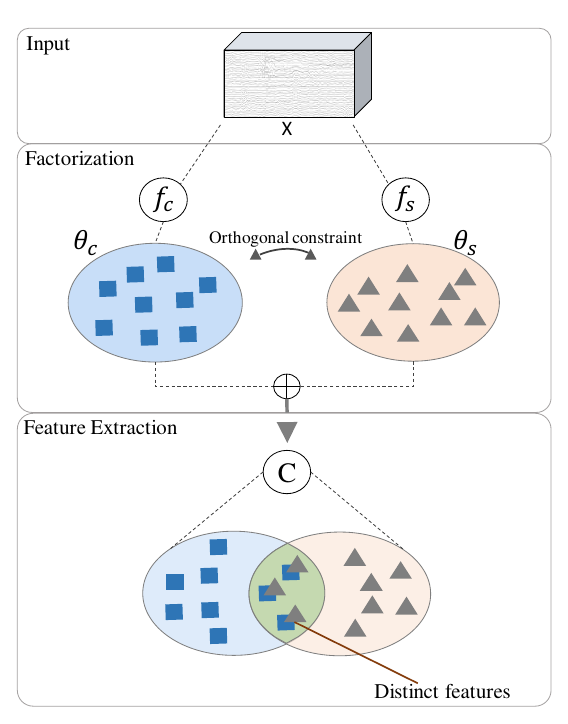}}
  \caption{Training scheme of the proposed method. The input signal are factorized through a class-common module $f_c$ and a class-specific module $f_s$. Factorization proceeds within orthogonal constraint, through which latent space $\theta_c$ and $\theta_s$ are separated. The features generated from the $f_c$ and $f_s$ are concatenated as an input of classifier $C$. Therefore, the $C$ enables consider decisive features from aggregated latent space.}
 \label{fig_paradigm}
\end{figure}

\section{Method}

\subsection{Overview}
The goal of this study is to divide features into two groups using factorization to extract distinct features under sparse condition. A sparse condition is defined as the absence of distinct spatial or temporal features for the different motor imagery classes. We designed two modules $f_c$ and $f_s$ to explicitly factorize EEG signals into class-common features $z_c$ and class-specific features $z_s$. The $z_c$ refers to common features of the EEG signals regardless of the class. The adversarial learning is applied for $f_c$ training. Both $z_c$ and $z_s$ are concatenated and fed into classifier $C$ for final prediction.

\subsection{Adversarial Learning}
Adversarial learning trains models to solve the $minmax$ optimization problem for the robustness of the models in several domains \cite{Fahimi, liu2016coupled, tzeng2017adversarial, zhong2020eeg, li2020novel}. Here, we use adversarial learning to enable $z_c$ to include common features of EEG signals, but not class-specific features. Unlike other studies \cite{ahmetouglu2021hierarchical, uelwer2021phase}, no generator is required because input data distributions are not utilized in this study. The training objective is to train $f_c$ for extraction of common features. The $f_c$ utilizes EEG signals $X = \{x_{i=1...N}\}$ to have mapping $\chi \rightarrow z_c$ where $X \in \chi$. Simultaneously, $f_c$ is trained to generate $z_c$ which can fool a discriminator $D$, while $D$ is trained to distinguish $z_c$ according to labels $K$ (real or fake). To generate fake features $z_c'$, we used the resting state class rather than using random noise $X'$. This is because $z_c$ and $z_c'$ should be similar, and the resting state is relatively similar to the input EEG signals than random noise. We defined the loss function as follows:

\begin{equation}
    L_{adv}(f_c, D, X^i, K^i)  = \underset{f_c}{min}\underset{D}{max} \sum_{i=0}^{N} K^i log(D(f_c(X^i))) 
\end{equation}
The class label $L$ is replaced $K$, thus $C$ and $f_s$ are not associated with this training. According to this, $D$ learns to maximize the probability of distinguishing correct $K$, while $f_c$ learns to generate $z_c$ and $z_c'$ similarly so that $D$ confuses to distinguish both features. 

\begin{table}[t!]
{\normalsize
\caption{Architecture of the proposed method. $\mathbb{F}$ and $\mathbb{T}$ denote feature and temporal dimension, respectively. ``[]'' and ``st.'' denote kernel size and stride, respectively.}
\renewcommand{\arraystretch}{1.2}
\resizebox{\columnwidth}{!}{%
\begin{tabular}{c|cccc} \hline
                                       & \textbf{$f_c$} & \textbf{$f_s$} & \textbf{$C$} & \textbf{$D$}      \\ \hline
\textbf{Input}                         & $X$ or $X'$        & $X$       & 1 $\times$ $\mathbb{F}$ $\times$ 2$\mathbb{T}$   & 1$\times$ $(\mathbb{F} \times \mathbb{T})$             \\ \hline
\multirow{4}{*}{\textbf{Hidden Layer}} & Conv. (1,40) [1,48]          & Conv. (1,40) [1,48]           & (5120, 2560)       & (5120, 2560) \\
                                       & Conv. (40,40) [24,1]          & Conv. (40,40) [24,1          & (2560, 1280)       & (2560, 1280) \\
                                       & Pool. (1, 68) st.=(1,14)          & Pool. (1, 68) st.=(1,14) & (1280, 640)       & (1280, 640) \\
                                       & Flatten          & Flatten          & (640, C)       & (640, 2)                \\ \hline
\textbf{Output}                        & 1 $\times$ $\mathbb{F}$ $\times$ $\mathbb{T}$       & 1$\times$ $\mathbb{F}$ $\times$ $\mathbb{T}$       & 1$\times$C        & 1$\times$K             \\ \hline
\textbf{Activation Function}           & \multicolumn{4}{c}{Exponential linear unit \cite{ELU}} \\ \hline       
\end{tabular}}}
\end{table}

\begin{table}[t!]
{\normalsize
\caption{Evaluation results of the existing methods and proposed method. Accuracy was calculated as the average of all folds. The reported accuracies are the average of all subjects. RF denotes random forest. The highest accuracy and the lowest standard deviation are bold}
\renewcommand{\arraystretch}{1}
\resizebox{\columnwidth}{!}{%
\begin{tabular}{c|cc|cc|cc} \hline
\backslashbox[40mm]{}   & \multicolumn{2}{c|}{Session 1} & \multicolumn{2}{c|}{Session 2} & \multicolumn{2}{c}{Session 3} \\ \hline
Model           & Acc           & std          & Acc           & std          & Acc           & std          \\\hline
CSP+LDA \cite{FBCSP}         & 0.21          & 0.02         & 0.26          & 0.03         & 0.20          & \textbf{0.01}         \\
CSP+RF \cite{qi2012random}         & 0.21          & 0.02         & 0.24          & \textbf{0.02}         & 0.19          & 0.04         \\
CSP+SVM \cite{FBCSP}        & 0.23          & \textbf{0.01}         & 0.25          & 0.04         & 0.21          & 0.02         \\
FBCSP \cite{FBCSP} & 0.26          & 0.03         & 0.28          & 0.05         & 0.23          & 0.03         \\
EEGNet \cite{eegnet}         & 0.45          & 0.04         & 0.43          & 0.07         & 0.37          & 0.05         \\
Shallow ConvNet \cite{deepconvnet} & 0.47          & 0.05         & 0.44          & 0.04         & 0.40          & 0.02         \\
Deep ConvNet \cite{deepconvnet}  & 0.45          & 0.02         & 0.42          & 0.05         & 0.38          & 0.03         \\
MCNN \cite{MCNN}    & 0.48          & 0.06         & 0.45          & 0.04         & 0.39          & 0.02         \\
Proposed Method & \textbf{0.52}          & 0.04         & \textbf{0.48}          & 0.06         & \textbf{0.45}          & 0.04     \\\hline   
\end{tabular}}}
\end{table}

\subsection{Architecture Configuration}
We designed $f_c$ and $f_s$ using convolution and pooling layers. Especially, $f_c$ and $f_s$ have no prediction layer because they conduct only feature extraction. $C$ and $D$ were designed based on multi-layer perceptron. From the $f_c$, the $z_c$ is fed into $C$ and $D$ as an input. In $C$, the features $z_c$ and $z_s$ are concatenated in temporal dimension to expand dimension for a series of convolutions. The last layer presents probabilities for each class using the softmax function. On the other hand, the $D$ flattens $z_c$ for linear regression. Through the regression layers, $D$ predicts the probability that the features are $z_c$ or $z_c'$. All layers include the exponential linear unit \cite{ELU} and drop out. Details of design choices are described in Table I.

\subsection{Training Scheme}
The $z_c$ is obtained by $f_c$ according to two objectives: $z_c$ is extracted similarly to $z_c'$; The $z_c$ contains classification-related representations. To this end, we use classification loss that is generated by $C$. The proposed method is designated to obtain mapping $\chi \rightarrow Y$ using $z_c$ and $z_s$. Specifically, $C$ learns mapping $z_c \oplus z_s \rightarrow Y$ by receiving features from $f_c$ and $f_s$. This is achieved through cross-entropy loss defined as

\begin{equation}
    L_{cls}(C, X^i, Y^i) = -\sum_{i=1}^{L} \hat{y^i} log(\sigma(C(X^i))) ,
\end{equation}
where $\hat{y^i}$ simply denotes class label. Through this, $f_c$, $f_s$ and $C$ share gradient of $L_{cls}$. Thus $f_c$ can consider $L_{adv}$ and $L_{cls}$ on the other hand, $f_s$ considers only $L_{cls}$ for training. The concept of the proposed training scheme is depicted in Fig. 2.

\subsection{Factorization}
To achieve the objective of factorization, $z_c$ and $z_s$ should have an orthogonal relationship. We introduced difference loss \cite{bousmalis2016domain} in order to divide latent space into two separate spaces. We consider $z_c$ and $z_s$ are different domain features hence the loss function is defined as:

\begin{equation}
    L_{diff}(f_c, f_s, X^i) = -\sum_{i=1}^{T} \parallel (f_c(X^i))^T f_s(X^i) \parallel ^2_F
\end{equation}
where $\parallel\cdot\parallel_F$ denotes Frobenius norm introduced in \cite{salzmann2010factorized}. Finally, the complete loss function is as follows:
 
\begin{equation}
    L_{all} = L_{cls} + L_{adv} + \lambda L_{diff}
\end{equation}
where $\lambda$ is regularization parameter for modulating the effect of $L_{diff}$. The flowchart of the factorization is depicted in Fig. 2. Additional parameters used in this study are described in the Experimental section.

\begin{figure*}[!t]
  \centerline{\includegraphics[scale = 0.95]{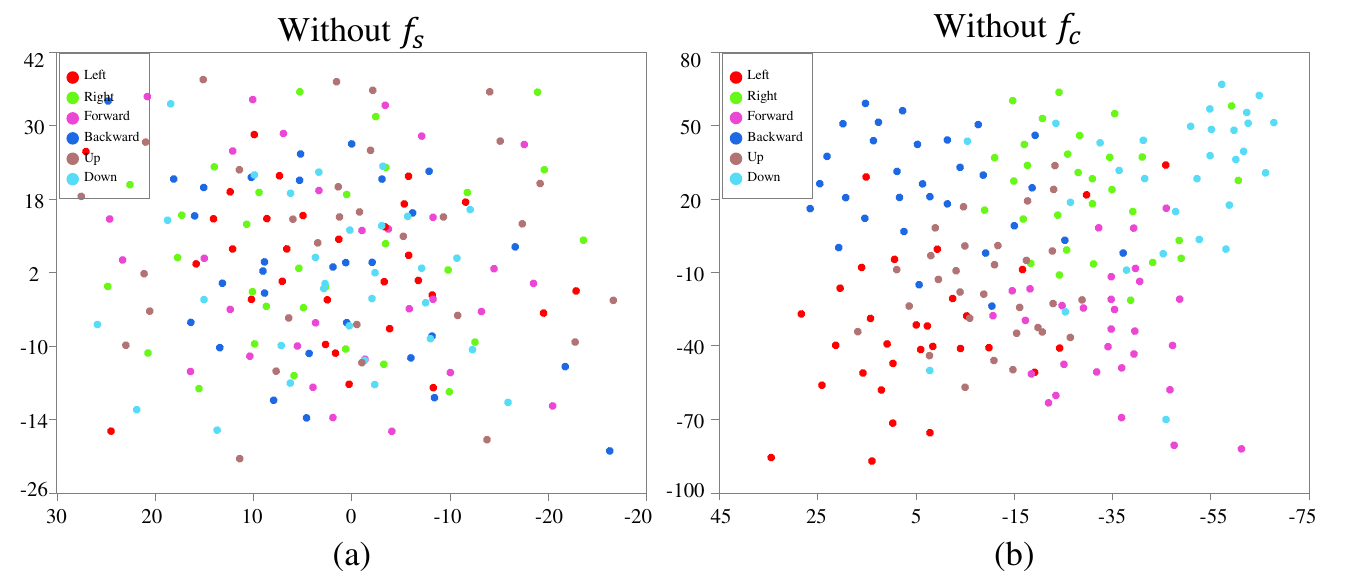}}
  \caption{Visualization of features acquired without each module using t-sne. (a) shows features mapped randomly without patterns. class-common module $f_c$ extracts features to fool the discriminator regardless of class. Therefore, it induces randomly mapped features of all classes. (b) visualizes features using only class-specific module $f_s$. Compared to (a), the features form clusters according to the classes hence $f_s$ is learned to perform classification. Two latent spaces are provided as an input of classifier $C$ therefore, the $C$ can consider two latent spaces that may contain discriminative representations.}
\end{figure*}

\begin{figure*}[!t]
  \centerline{\includegraphics[scale = 0.95]{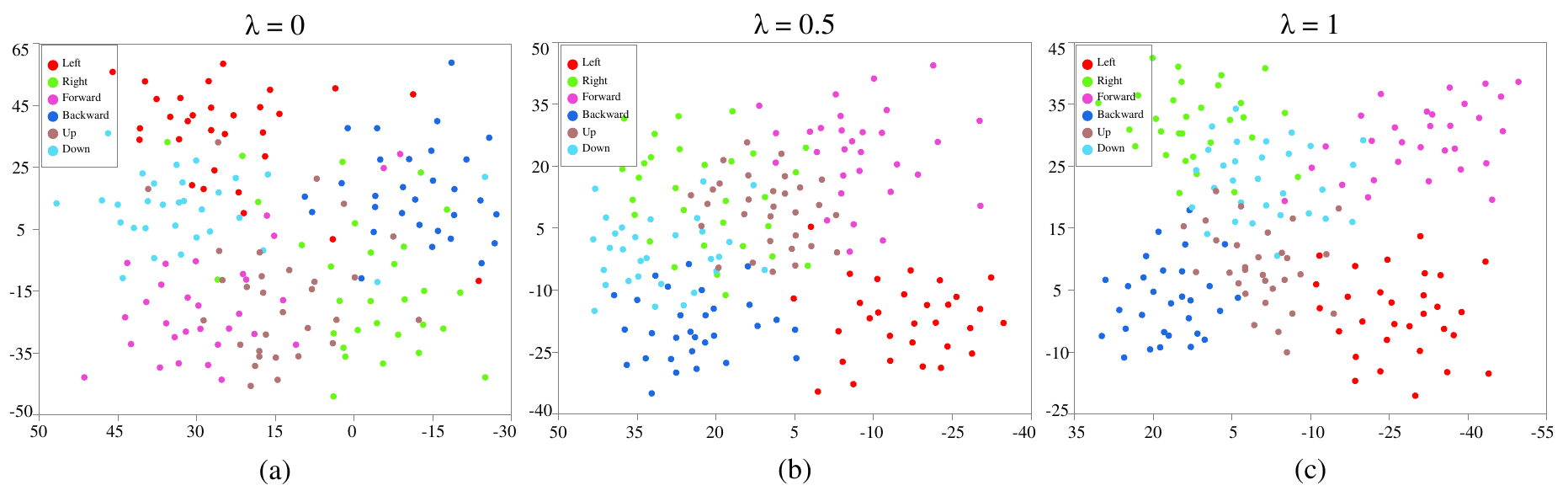}}
  \caption{Visualization of features by the classifier. The difference loss $L_{diff}$ was introduced for orthogonal constraints with regularization factor $\lambda$. (a) shows the features without the effect of $L_{diff}$. The distribution of features is similar to those of features shown in Fig. 3(b). They share the latent space with other classes. In (b), the features are clustered by class. However, the classes ``up'', ``right'', and ``down'' still share some of the latent space. (c) shows that features form relatively clear clusters in the latent space. Although they share some latent spaces with the ``down'' and ``backward'' classes, the features are well lumped together at the center of each cluster.}
\end{figure*}

\section{Experiments}
We evaluated the proposed model on the single-arm movements imagery dataset \cite{jeong2020multimodal}. We choose 7 classes (``left'', ``right'', ``up'', ``down'', ``forward'', ``backward'', and ``resting state'') among 12 classes and used 6 classes without resting state for evaluation. Note that randomly selected trials of resting state were utilized for adversarial learning only. Each class contains 50 trials thus organized dataset yields 300 trials for each subject. The dataset is composed of 3 recording sessions, all sessions were used for evaluation in a subject-dependent manner. Performance was measured by the average of all folds. We applied 5-fold cross-validation yielding 30 trials of the training set and 10 trials of validation and test set for each fold. As a data augmentation technique, data cropping was introduced with a 100 ms size of sliding window introduced in \cite{deepconvnet}. To avoid multiple outputs of the prediction layer, all crops were averaged for making a single prediction. The training epoch was 400 epochs and the model weights reporting the lowest validation loss were stored after 200 epoch. We applied AdamW optimizer \cite{AdamW} and the learning rate was 0.001 with 0.01 of weight decay. The value of $\lambda$ was 1. The experiment was conducted on an Intel 3.60 Core i7 9700 K CPU with 64 GB of RAM, two NVIDIA TITAN V GPUs, and Python version 3.9 with PyTorch version 1.9.

\begin{table}[t!]
\centering
{\normalsize
\caption{Experimental results of ablation study. Regardless of the session, performances decreased when each module was independently used. In particular, when only $f_c$ was used for training, the classification performance decreased significantly. Without $f_c$, a pipeline of the proposed method is the same as other existing methods. However, jointly use of both modules derived performance improvement}
\renewcommand{\arraystretch}{1}
\begin{tabular}{c|cc|cc|cc} \hline
\backslashbox[20mm] {}     & \multicolumn{2}{c|}{Session 1} & \multicolumn{2}{c|}{Session 2} & \multicolumn{2}{c}{Session 3} \\ \hline
Model    & Acc           & std          & Acc           & std          & Acc           & std          \\\hline
W/o $f_c$ & 0.25          & 0.01         & 0.26          & 0.03         & 0.22          & 0.02         \\
W/o $f_s$ & 0.45          & 0.03         & 0.44          & 0.02         & 0.39          & 0.02         \\
Both     & 0.52          & 0.04         & 0.48          & 0.03         & 0.45          & 0.04 \\\hline        
\end{tabular}}
\end{table}

\begin{table}[t!]
\centering
{\normalsize
\caption{Evaluation on session 1 by changing the value of regularization factor $\lambda$. It modulates the influence of the difference loss, in this experiment, the performance improved as it increased. It can be inferred that the orthogonality of the latent space improves performance.}
\renewcommand{\arraystretch}{1}
\begin{tabular}{c|cccc} \hline
\backslashbox[10mm] {}      & $\lambda$ = 0 & $\lambda$ = 0.5 & $\lambda$ = 1 \\ \hline
Avg. & 0.47  & 0.50    & 0.52  \\
std. & 0.02  & 0.04    & 0.04  \\ \hline
\end{tabular}}
\end{table}

\subsection{Results}
Model performances on \cite{jeong2020multimodal} were lower than the reported performances \cite{eegnet, deepconvnet, MCNN, FBCSP} according to Table II. This can be inferred that the dataset contains imagery tasks in which distinct features are sparse, as we assume. Specifically, CSP-based methods focusing on spatial patterns showed approximately 0.23 points. Since it consists of movements within a single-arm, the origins of the EEG signals are small brain region according to \cite{decety1996neurophysiological, michel2019eeg}. It is difficult to extract distinct spatial features thus, CSP is disadvantageous for this type of dataset. However, CNN-based methods show relatively robust to sparse condition. The average performance of the methods is about 0.43 points. EEGNet showed efficient performance even with a small number of parameters. Shallow ConvNet recorded the competitive performance, as it is known to be specialized in MI classification. It is the highest accuracy among methods of having a single architecture. MCNN consisted of multiple CNN architecture for multi-view feature extraction. According to Table II, it was robust to the sparse condition. The proposed method showed the highest performance in this experiment. It is inferred that factorization approach enables model to extract distinct features for learning under sparse condition.

\subsection{Ablation Study}
\subsubsection{Comparison of $f_c$ and $f_s$}
Firstly, an experiment was conducted to confirm whether the $f_c$ extracts common features regardless of class. Using only $f_c$ showed twice as low performance as reported according to Table III. Although $f_c$ was trained for classification using $L_{cls}$, adversarial learning rather interferes with training efficiency. The $f_c$ was designed to extract features regardless of class, hence $z_c$ contains not enough class-discriminative features. Furthermore, no distinct patterns or clusters were shown by features obtained by $f_c$ as depicted in Fig. 3(a). On the other hand, when using only $f_s$, it showed similar performance to existing methods because it has the same training scheme as other methods. The features extracted by $f_s$ relatively form clusters for each class as shown in Fig. 3(b). It can be inferred that distinct features were selected for each class. However, the proposed method showed the highest accuracy using two modules jointly. It means that the concatenation of the features induced significant features in classification. All results are described in Table III.

%%%% 여기서부터 시작
\subsubsection{Comparison of $\lambda$ value}
We evaluated model performance by changing the values of $\lambda$, and the results are shown in Table IV. The $L_{diff}$ was introduced to disjoint $z_c$ and $z_s$ by establishing an orthogonal relationship. Performance was improved as the influence of $L_{diff}$ increased according to Table IV. Thus, it demonstrated that orthogonal relationship could improve classification performance. This can also be confirmed through visualization. Fig. 4 provides visualization of features obtained by $C$. When the $L_{diff}$ was not influential ($\lambda$ = 0), the features showed a similar distribution with Fig. 3(b) rather than distinguishing clusters. In Fig 4(b), a clearer clusters were formed, specifically, the features of ``left'' class are gathered around the cluster. However, numerous features share latent space, especially ``Up'' class is totally not distinguishable. Finally, features formed clusters by class densely on the center of the cluster in Fig. 4(c). Some features still share the space, but it shows the clearest clusters. Therefore, we confirmed that the $L_{diff}$ enabled the model to extract distinct and decisive features in performance according to results and visualizations. In summary, we demonstrated that the designed modules were trained to obtain two separate latent spaces through the factorization. In addition, jointly training of separate latent spaces derived performance improvement under sparse condition.

\section{Discussion}
\subsection{Sparse Condition}
The goal of this study is to improve classification performance by sufficiently extracting decisive features under sparse condition. Firstly, we considered the definition of sparse condition. The brain region where the EEG signals are generated should be a small area, thus spatial characteristics of the imagery tasks should be similar. Publicly opened MI BCI datasets \cite{competition, ofner2017upper} are mainly composed of imagery tasks that separate brain regions (e.g. foot, tongue, left and right hand) \cite{decety1996neurophysiological, michel2019eeg}. Jeong \textit{et al.} \cite{jeong2020multimodal} proposed single-arm movement imagery tasks such as 6 directions of arm reaching, 3 different grasping. Since all subjects are right-handed, the mainly activated brain region would be the left sensorimotor cortex. Specifically, according to functional brain mapping, meaningful EEG signals are intensively generated only in the middle region of the sensorimotor cortex \cite{crone1998functional}. To introduce the sparse condition, we selected 6 directions of arm reaching classes because they share the same brain region. Based on provided experimental protocol, we conjecture that it would be fewer distinct temporal and spatial features between classes. Through the experiments, we confirmed that existing methods showed unsatisfactory performance hence conventional feature extraction is not efficient under sparse condition. However further studies are needed to clearly explore sparse condition in various ways because of the lack of ground truth of EEG signals.
\subsection{Class-common Features}
We designed $z_c$ to include unrelated features to the classes. Therefore, it may include common representation of the EEG signals. It is not known whether common information is spatial, temporal, or contains all. In the spatial aspect, features would include information of middle region of the left sensorimotor cortex. On the one hand, in the temporal aspect, features may contain information ``reaching arm'' over time except for the direction according to the experimental protocol. Ablation study showed that using only $z_c$ was not efficient for classification because it is class-unrelated features as shown in Table III and Fig. 3(a). Nevertheless, it helps the model extract more distinct features when jointly used with $z_s$. This means that $z_c$ broadens the view of $C$ to consider implicit representations in latent space. Indeed $C$ acquired features that formed clusters for each class more clearly with the $L_{diff}$ depicted as in Fig. 4.

\subsection{Class-specific Features}
Class-specific features, $z_s$ in this study, are the same as those produced in commonly used training. In Fig. 3(b), the features relatively appear to cluster in the latent space compared to Fig. 3(a) although some features are apparent outliers. In addition, when only $z_s$ was used for classification, the performance was similar to those of existing methods. However, when the classifier considered both $z_s$ and $z_c$, it acquired the most distinctive representations and showed the highest performance. As a result, we demonstrated that $z_c$ helps $C$ acquire more decisive features for classification than under sparse condition.

\subsection{Orthogonality in EEG Features}
The $L_{diff}$ constrains $z_c$ and $z_s$ to be orthogonal to each other. Our approach is that the $z_c$ and $z_s$ are disjoint with each other so that allowing $C$ to consider implicit representations in a wider latent space through concatenation. Thus we conjecture that $z_c$ and $z_s$ are different domain features. Thus $L_{diff}$ was introduced to minimize the latent space shared by the two features. As shown in Fig 3(a), $L_{diff}$, the features from $C$ form clusters with numerous outliers in the latent space. As the influence of $L_{diff}$ increases, classification accuracies improved, and features also form clear clusters. Through this, we confirmed that orthogonal constraint allows $C$ to use more abundant features from separate latent spaces.

% 분류기에게 직접적으로 풍부한 피쳐를 뽑게하는게 아니라, 피쳐를 뽑을 수 있는 넓은 시야를 주고 그중에서 잘 특징을 뽑으면 분류 성능이 높아진다 라는 가정/// 클래스들끼리 겹치는 양상이 있음을 확인

\section{Conclusions}
In this paper, we defined sparse condition propose a factorization approach that integrates latent space of class-common and class-specific features. Under the sparse condition, existing feature extraction methods face difficulty extracting distinct features, and their performance also decreases. To solve this problem, we introduced a training strategy that factorizes EEG signals in order to obtain distinct features under sparse condition. To this end, we applied a regime of adversarial learning to extract class-common features. The proposed method is trained to deceive the discriminator by extracting features regardless of class. A training strategy is to factorize EEG signals into two types of features so that it allows the classifier to consider the discriminative features from the two different latent spaces. In order to minimize the shared latent spaces by the features, orthogonal constraints were introduced based on the difference loss function. Experimental results demonstrated that the factorization achieved performance improvement in single-arm MI classification accuracy by integrating class-common and class-specific features. Although the proposed method achieved performance improvement, it is necessary to investigate what information class-common features contain and what effect the combination of latent spaces plays in obtaining distinct features. Therefore, our future works are to explore the issues to clearly prove it.

\section{Acknowledgements}
This work was partly supported by Institute of Information \& Communications Technology Planning \& Evaluation (IITP) grant funded by the Korea government (MSIT) (No. 2017-0-00432, Development of Non-Invasive Integrated BCI SW Platform to Control Home Appliances and External Devices by User’s Thought via AR/VR Interface; No. 2017-0-00451, Development of BCI based Brain and Cognitive Computing Technology for Recognizing User’s Intentions using Deep Learning; No. 2019-0-00079, Artificial Intelligence Graduate School Program, Korea University).

% 스파시오-템퍼럴 희소 조건 정의하기 

\bibliographystyle{./IEEEtran}
\bibliography{./REFERENCES}

% Generated by IEEEtran.bst, version: 1.12 (2007/01/11)
\begin{thebibliography}{10}
\providecommand{\url}[1]{#1}
\csname url@samestyle\endcsname
\providecommand{\newblock}{\relax}
\providecommand{\bibinfo}[2]{#2}
\providecommand{\BIBentrySTDinterwordspacing}{\spaceskip=0pt\relax}
\providecommand{\BIBentryALTinterwordstretchfactor}{4}
\providecommand{\BIBentryALTinterwordspacing}{\spaceskip=\fontdimen2\font plus
\BIBentryALTinterwordstretchfactor\fontdimen3\font minus
  \fontdimen4\font\relax}
\providecommand{\BIBforeignlanguage}[2]{{%
\expandafter\ifx\csname l@#1\endcsname\relax
\typeout{** WARNING: IEEEtran.bst: No hyphenation pattern has been}%
\typeout{** loaded for the language `#1'. Using the pattern for}%
\typeout{** the default language instead.}%
\else
\language=\csname l@#1\endcsname
\fi
#2}}
\providecommand{\BIBdecl}{\relax}
\BIBdecl

\bibitem{chen2021points}
L.~Chen, T.~Yang, X.~Zhang, W.~Zhang, and J.~Sun, ``Points as queries: Weakly
  semi-supervised object detection by points,'' in \emph{Proc. IEEE Comput.
  Soc. Conf. Comput. Vis. Pattern Recognit. (CVPR)}, 2021, pp. 8823--8832.

\bibitem{lee1997new}
S.-W. Lee and H.-H. Song, ``A new recurrent neural-network architecture for
  visual pattern recognition,'' \emph{IEEE trans. Neural Netw.}, vol.~8, no.~2,
  pp. 331--340, 1997.

\bibitem{kao2021activity}
P.~Y. Kao, Y.-J. Lei, C.-H. Chang, C.-S. Chen, M.-S. Lee, and Y.-P. Hung,
  ``Activity recognition using first-person-view cameras based on sparse
  optical flows,'' in \emph{Proc. Int. Conf. Pattern Recognit. (ICPR)}.\hskip
  1em plus 0.5em minus 0.4em\relax IEEE, 2021, pp. 81--86.

\bibitem{lee2022human}
D.-G. Lee and S.-W. Lee, ``Human interaction recognition framework based on
  interacting body part attention,'' \emph{Pattern Recognit.}, vol. 128, p.
  108645, 2022.

\bibitem{lee2021voicemixer}
S.-H. Lee, J.-H. Kim, H.~Chung, and S.-W. Lee, ``Voicemixer: {A}dversarial
  {V}oice {S}tyle {M}ixup,'' in \emph{Thirty-Fifth Conf. Adv. Neural Inf.
  Process Syst.}, Dec, 2021.

\bibitem{liu2021audio}
H.~Liu, W.~Xu, and B.~Yang, ``Audio-visual speech recognition using a two-step
  feature fusion strategy,'' in \emph{Proc. Int. Conf. Pattern Recognit.
  (ICPR)}.\hskip 1em plus 0.5em minus 0.4em\relax IEEE, 2021, pp. 1896--1903.

\bibitem{lee2022duration}
S.-H. Lee, H.-R. Noh, W.-J. Nam, and S.-W. Lee, ``Duration controllable voice
  conversion via phoneme-based information bottleneck,'' \emph{IEEE/ACM Trans.
  Audio, Speech, Language Process.}, vol.~30, pp. 1173--1183, 2022.

\bibitem{liu2021mutual}
H.~Liu, Y.~Wang, and B.~Yang, ``Mutual alignment between audiovisual features
  for end-to-end audiovisual speech recognition,'' in \emph{Proc. Int. Conf.
  Pattern Recognit. (ICPR)}.\hskip 1em plus 0.5em minus 0.4em\relax IEEE, 2021,
  pp. 5348--5353.

\bibitem{Lee_Yoon_Noh_Kim_Lee_2021}
S.-H. Lee, H.-W. Yoon, H.-R. Noh, J.-H. Kim, and S.-W. Lee,
  ``Multi-spectro{GAN}: {H}igh-{D}iversity and {H}igh-{F}idelity {S}pectrogram
  {G}eneration with {A}dversarial {S}tyle {C}ombination for {S}peech
  {S}ynthesis,'' in \emph{Proc. AAAI Conf. Artif. Intell.}, vol.~35, no.~14,
  New York, USA, Feb, 2021, pp. 13\,198--13\,206.

\bibitem{devlin2018bert}
J.~Devlin, M.-W. Chang, K.~Lee, and K.~Toutanova, ``Bert: Pre-training of deep
  bidirectional transformers for language understanding,'' \emph{arXiv preprint
  arXiv:1810.04805}, 2018.

\bibitem{lee2021korealbert}
H.~Lee, J.~Yoon, B.~Hwang, S.~Joe, S.~Min, and Y.~Gwon, ``Korealbert:
  Pretraining a lite bert model for korean language understanding,'' in
  \emph{Proc. Int. Conf. Pattern Recognit. (ICPR)}.\hskip 1em plus 0.5em minus
  0.4em\relax IEEE, 2021, pp. 5551--5557.

\bibitem{yang2021pose}
Z.~Yang, A.~Kay, Y.~Li, W.~Cross, and J.~Luo, ``Pose-based body language
  recognition for emotion and psychiatric symptom interpretation,'' in
  \emph{Proc. Int. Conf. Pattern Recognit. (ICPR)}.\hskip 1em plus 0.5em minus
  0.4em\relax IEEE, 2021, pp. 294--301.

\bibitem{lee2022quantifying}
M.~Lee, L.~R. Sanz, A.~Barra, A.~Wolff, J.~O. Nieminen, M.~Boly, M.~Rosanova,
  S.~Casarotto, O.~Bodart, J.~Annen \emph{et~al.}, ``Quantifying arousal and
  awareness in altered states of consciousness using interpretable deep
  learning,'' \emph{Nat. Commun.}, vol.~13, no.~1, pp. 1--14, 2022.

\bibitem{cho2021neurograsp}
J.-H. Cho, J.-H. Jeong, and S.-W. Lee, ``Neurograsp: Real-time eeg
  classification of high-level motor imagery tasks using a dual-stage deep
  learning framework,'' \emph{IEEE Trans. Cybern.}, 2021.

\bibitem{won2017motion}
D.-O. Won, H.-J. Hwang, D.-M. Kim, K.-R. M{\"u}ller, and S.-W. Lee,
  ``Motion-based rapid serial visual presentation for gaze-independent
  brain-computer interfaces,'' \emph{IEEE Trans. Neural Syst. Rehabil. Eng.},
  vol.~26, no.~2, pp. 334--343, 2017.

\bibitem{wheelchair}
K.-T. Kim, H.-I. Suk, and S.-W. Lee, ``Commanding a brain-controlled wheelchair
  using steady-state somatosensory evoked potentials,'' \emph{IEEE Trans.
  Neural Syst. Rehabil. Eng.}, vol.~26, no.~3, pp. 654--665, 2016.

\bibitem{chen2016high}
Y.~Chen, A.~D. Atnafu, I.~Schlattner, W.~T.~a. Weldtsadik, and S.~Fazli, ``A
  high-security {EEG}-based login system with {RSVP} stimuli and dry
  electrodes,'' \emph{IEEE Trans. Inf. Forensics Secur.}, vol.~11, no.~12, pp.
  2635--2647, 2016.

\bibitem{suk2014predicting}
H.-I. Suk, S.~Fazli, J.~Mehnert, K.-R. M{\"u}ller, and S.-W. Lee, ``Predicting
  bci subject performance using probabilistic spatio-temporal filters,''
  \emph{PloS one}, vol.~9, no.~2, p. e87056, 2014.

\bibitem{lee2022motor}
D.-Y. Lee, J.-H. Jeong, B.-H. Lee, and S.-W. Lee, ``Motor imagery
  classification using inter-task transfer learning via a channel-wise
  variational autoencoder-based convolutional neural network,'' \emph{IEEE
  Trans. Neural Syst. Rehab. Eng.}, vol.~30, pp. 226--237, 2022.

\bibitem{hill2012recording}
N.~J. Hill, D.~Gupta, P.~Brunner, A.~Gunduz, M.~A. Adamo, A.~Ritaccio, and
  G.~Schalk, ``Recording human electrocorticographic (ecog) signals for
  neuroscientific research and real-time functional cortical mapping,''
  \emph{J. Vis. Exp.}, no.~64, p. e3993, 2012.

\bibitem{lee2015subject}
M.-H. Lee, S.~Fazli, J.~Mehnert, and S.-W. Lee, ``Subject-dependent
  classification for robust idle state detection using multi-modal neuroimaging
  and data-fusion techniques in {BCI},'' \emph{Pattern Recognit.}, vol.~48,
  no.~8, pp. 2725--2737, 2015.

\bibitem{mcavinue2008measuring}
L.~P. McAvinue and I.~H. Robertson, ``Measuring motor imagery ability: a
  review,'' \emph{Eur. J. Cogn. Psychol.}, vol.~20, no.~2, pp. 232--251, 2008.

\bibitem{sousa2017pure}
T.~Sousa, C.~Amaral, J.~Andrade, G.~Pires, U.~J. Nunes, and M.~Castelo-Branco,
  ``Pure visual imagery as a potential approach to achieve three classes of
  control for implementation of {BCI} in non-motor disorders,'' \emph{J. Neural
  Eng.}, vol.~14, no.~4, p. 046026, 2017.

\bibitem{DaSalla}
C.~S. DaSalla, H.~Kambara, M.~Sato, and Y.~Koike, ``Single-trial classification
  of vowel speech imagery using common spatial patterns,'' \emph{Neural Netw.},
  vol.~22, no.~9, pp. 1334--1339, Nov. 2009.

\bibitem{zhang2017hybrid}
Y.~Zhang, H.~Zhang, X.~Chen, S.-W. Lee, and D.~Shen, ``{Hybrid High-Order
  Functional Connectivity Networks Using Resting-state Functional MRI for Mild
  Cognitive Impairment Diagnosis},'' \emph{Sci. Rep.}, vol.~7, no.~1, pp.
  1--15, 2017.

\bibitem{lee2020classification}
B.-H. Lee, J.-H. Jeong, K.-H. Shim, and S.-W. Lee, ``Classification of
  high-dimensional motor imagery tasks based on an end-to-end role assigned
  convolutional neural network,'' in \emph{IEEE Int. Conf. Acoust. Speech
  Signal Process. (ICASSP)}.\hskip 1em plus 0.5em minus 0.4em\relax IEEE, 2020,
  pp. 1359--1363.

\bibitem{bci3}
N.-S. Kwak, K.-R. M{\"u}ller, and S.-W. Lee, ``A lower limb exoskeleton control
  system based on steady state visual evoked potentials,'' \emph{J. Neural.
  Eng.}, vol.~12, no.~5, p. 056009, 2015.

\bibitem{decety1996neurophysiological}
J.~Decety, ``The neurophysiological basis of motor imagery,'' \emph{Behav.
  Brain Res.}, vol.~77, no. 1-2, pp. 45--52, 1996.

\bibitem{jeong2020decoding}
J.-H. Jeong, N.-S. Kwak, C.~Guan, and S.-W. Lee, ``{Decoding movement-related
  cortical potentials based on subject-dependent and section-wise spectral
  filtering},'' \emph{IEEE Trans. Neural Syst. Rehab. Eng.}, vol.~28, no.~3,
  pp. 687--698, 2020.

\bibitem{kwon2020decoding}
B.-H. Kwon, J.-H. Jeong, J.-H. Cho, and S.-W. Lee, ``{Decoding of Intuitive
  Visual Motion Imagery Using Convolutional Neural Network under 3D-BCI
  Training Environment},'' in \emph{Int. Conf. Sys. Man, and Cybern. (SMC)},
  Toronto, Canada, Oct, 2020, pp. 2966--2971.

\bibitem{dasalla2009single}
C.~S. DaSalla, H.~Kambara, M.~Sato, and Y.~Koike, ``Single-trial classification
  of vowel speech imagery using common spatial patterns,'' \emph{Neural Netw.},
  vol.~22, no.~9, pp. 1334--1339, 2009.

\bibitem{Nguyen}
C.~H. Nguyen, G.~K. Karavas, and P.~Artemiadis, ``{Inferring imagined speech
  using {EEG} signals: a new approach using {R}iemannian manifold features},''
  \emph{J. Neural Eng.}, vol.~15, no.~1, p. 016002, Dec. 2017.

\bibitem{jeong2020multimodal}
J.-H. Jeong, J.-H. Cho, K.-H. Shim, B.-H. Kwon, B.-H. Lee, D.-Y. Lee, D.-H.
  Lee, and S.-W. Lee, ``Multimodal signal dataset for 11 intuitive movement
  tasks from single upper extremity during multiple recording sessions,''
  \emph{GigaScience}, vol.~9, no.~10, p. giaa098, Oct. 2020.

\bibitem{competition}
M.~Tangermann, K.-R. M{\"u}ller, A.~Aertsen, N.~Birbaumer, C.~Braun,
  C.~Brunner, R.~Leeb, C.~Mehring, K.~J. Miller, G.~Mueller-Putz \emph{et~al.},
  ``Review of the {BCI} competition {IV},'' \emph{Front. Neurosci.}, vol.~6,
  p.~55, 2012.

\bibitem{lee2021decoding}
D.-Y. Lee, M.~Lee, and S.-W. Lee, ``Decoding imagined speech based on deep
  metric learning for intuitive bci communication,'' \emph{IEEE Trans. Neural
  Syst. Rehab. Eng.}, vol.~29, pp. 1363--1374, 2021.

\bibitem{eegnet}
V.~J. Lawhern, A.~J. Solon, N.~R. Waytowich, S.~M. Gordon, C.~P. Hung, and
  B.~J. Lance, ``E{EGN}et: a compact convolutional neural network for eeg-based
  brain--computer interfaces,'' \emph{J. Neural Eng.}, vol.~15, no.~5, p.
  056013, 2018.

\bibitem{deepconvnet}
R.~T. Schirrmeister, J.~T. Springenberg, L.~D.~J. Fiederer, M.~Glasstetter,
  K.~Eggensperger, M.~Tangermann, F.~Hutter, W.~Burgard, and T.~Ball, ``Deep
  learning with convolutional neural networks for {EEG} decoding and
  visualization,'' \emph{Hum. Brain Mapp.}, vol.~38, no.~11, pp. 5391--5420,
  2017.

\bibitem{deeplearning}
A.~M. Azab, L.~Mihaylova, K.~K. Ang, and M.~Arvaneh, ``Weighted transfer
  learning for improving motor imagery-based brain--computer interface,''
  \emph{IEEE Trans. Neural Syst. Rehabil. Eng.}, vol.~27, no.~7, pp.
  1352--1359, 2019.

\bibitem{OYKwon}
O.-Y. Kwon, M.-H. Lee, C.~Guan, and S.-W. Lee, ``Subject-independent
  brain-computer interfaces based on deep convolutional neural networks,''
  \emph{IEEE Trans. Neural Netw. Learn. Syst.}, vol.~31, no.~10, pp.
  3839--3852, 2020.

\bibitem{deeplearning2}
K.~Venkatachalam, A.~Devipriya, J.~Maniraj, M.~Sivaram, A.~Ambikapathy, and
  S.~A. Iraj, ``A novel method of motor imagery classification using {EEG}
  signal,'' \emph{Artif. Intell. Med.}, vol. 103, p. 101787, 2020.

\bibitem{JHJeong}
J.-H. Jeong, K.-H. Shim, D.-J. Kim, and S.-W. Lee, ``Brain-controlled robotic
  arm system based on multi-directional {CNN}-{B}i{LSTM} network using {EEG}
  signals,'' \emph{IEEE Trans. Neural Syst. Rehabil. Eng.}, vol.~28, no.~5, pp.
  1226--1238, 2020.

\bibitem{MCNN}
S.~U. Amin, M.~Alsulaiman, G.~Muhammad, M.~A. Bencherif, and M.~S. Hossain,
  ``Multilevel weighted feature fusion using convolutional neural networks for
  {EEG} motor imagery classification,'' \emph{IEEE Access}, vol.~7, pp.
  18\,940--18\,950, 2019.

\bibitem{Lu}
N.~Lu, T.~Li, X.~Ren, and H.~Miao, ``A deep learning scheme for motor imagery
  classification based on restricted {B}oltzmann machines,'' \emph{IEEE Trans.
  Neural Syst. Rehabil. Eng.}, vol.~25, no.~6, pp. 566--576, 2016.

\bibitem{FBCSP}
K.~K. Ang, Z.~Y. Chin, H.~Zhang, and C.~Guan, ``Filter bank common spatial
  pattern ({FBCSP}) in brain-computer interface,'' in \emph{Proc. Int. Jt.
  Conf. Neural Netw.}\hskip 1em plus 0.5em minus 0.4em\relax IEEE, 2008, pp.
  2390--2397.

\bibitem{sakhavi}
S.~Sakhavi, C.~Guan, and S.~Yan, ``Learning temporal information for
  brain-computer interface using convolutional neural networks,'' \emph{IEEE
  Trans. Neural Netw. Learn. Syst.}, vol.~29, no.~11, pp. 5619--5629, 2018.

\bibitem{burle2015spatial}
B.~Burle, L.~Spieser, C.~Roger, L.~Casini, T.~Hasbroucq, and F.~Vidal,
  ``Spatial and temporal resolutions of {EEG}: {I}s it really black and white?
  {A} scalp current density view,'' \emph{Int. J. Psychophysiol.}, vol.~97,
  no.~3, pp. 210--220, 2015.

\bibitem{Torres-Garcia}
A.~A. Torres-Garc{\'\i}a, C.~A. Reyes-Garc{\'\i}a, L.~Villase{\~n}or-Pineda,
  and G.~Garc{\'\i}a-Aguilar, ``{Implementing a fuzzy inference system in a
  multi-objective EEG channel selection model for imagined speech
  classification},'' \emph{Expert Syst. Appl.}, vol.~59, pp. 1--12, Oct. 2016.

\bibitem{lee2003pattern}
S.-W. Lee and A.~Verri, \emph{{Pattern Recognition with Support Vector
  Machines: First International Workshop, SVM 2002, Niagara Falls, Canada,
  August 10, 2002. Proceedings}}.\hskip 1em plus 0.5em minus 0.4em\relax
  Springer, 2003, vol. 2388.

\bibitem{Fahimi}
F.~Fahimi, S.~Dosen, K.~K. Ang, N.~Mrachacz-Kersting, and C.~Guan, ``Generative
  adversarial networks-based data augmentation for brain-computer interface,''
  \emph{IEEE Trans. Neural Netw. Learn. Syst.}, 2020.

\bibitem{liu2016coupled}
M.-Y. Liu and O.~Tuzel, ``Coupled generative adversarial networks,'' \emph{Adv.
  Neural Inf. Process. Syst. (NIPS)}, vol.~29, pp. 469--477, 2016.

\bibitem{tzeng2017adversarial}
E.~Tzeng, J.~Hoffman, K.~Saenko, and T.~Darrell, ``Adversarial discriminative
  domain adaptation,'' in \emph{Proc. IEEE Comput. Conf. Comput. Vis. Pattern
  Recognit. (CVPR)}, 2017, pp. 7167--7176.

\bibitem{zhong2020eeg}
P.~Zhong, D.~Wang, and C.~Miao, ``Eeg-based emotion recognition using
  regularized graph neural networks,'' \emph{IEEE Trans. Affect. Comput.},
  2020.

\bibitem{li2020novel}
Y.~Li, L.~Wang, W.~Zheng, Y.~Zong, L.~Qi, Z.~Cui, T.~Zhang, and T.~Song, ``A
  novel bi-hemispheric discrepancy model for eeg emotion recognition,''
  \emph{IEEE Trans. Cogn. Devel. Syst.}, vol.~13, no.~2, pp. 354--367, 2020.

\bibitem{ahmetouglu2021hierarchical}
A.~Ahmeto{\u{g}}lu and E.~Alpayd{\i}n, ``Hierarchical mixtures of generators
  for adversarial learning,'' in \emph{Proc. Int. Conf. Pattern Recognit.
  (ICPR)}.\hskip 1em plus 0.5em minus 0.4em\relax IEEE, 2021, pp. 316--323.

\bibitem{uelwer2021phase}
T.~Uelwer, A.~Oberstra$\beta$, and S.~Harmeling, ``Phase retrieval using
  conditional generative adversarial networks,'' in \emph{Proc. Int. Conf.
  Pattern Recognit. (ICPR)}.\hskip 1em plus 0.5em minus 0.4em\relax IEEE, 2021,
  pp. 731--738.

\bibitem{ELU}
D.-A. Clevert, T.~Unterthiner, and S.~Hochreiter, ``Fast and accurate deep
  network learning by exponential linear units (elus),'' \emph{arXiv preprint
  arXiv:1511.07289}, 2015.

\bibitem{qi2012random}
Y.~Qi, ``Random forest for bioinformatics,'' in \emph{Ensem. mach. learn.},
  2012, pp. 307--323.

\bibitem{bousmalis2016domain}
K.~Bousmalis, G.~Trigeorgis, N.~Silberman, D.~Krishnan, and D.~Erhan, ``Domain
  separation networks,'' \emph{Adv. Neural Inf. Process. Syst.}, vol.~29, pp.
  343--351, 2016.

\bibitem{salzmann2010factorized}
M.~Salzmann, C.~H. Ek, R.~Urtasun, and T.~Darrell, ``Factorized orthogonal
  latent spaces,'' in \emph{Proc. Thirteenth Int. Conf. Artif. Intell.
  Stat.}\hskip 1em plus 0.5em minus 0.4em\relax JMLR Workshop and Conference
  Proceedings, 2010, pp. 701--708.

\bibitem{AdamW}
I.~Loshchilov and F.~Hutter, ``Decoupled weight decay regularization,''
  \emph{arXiv preprint arXiv:1711.05101}, 2017.

\bibitem{michel2019eeg}
C.~M. Michel and D.~Brunet, ``{EEG} source imaging: a practical review of the
  analysis steps,'' \emph{Front. Neurol.}, vol.~10, p. 325, 2019.

\bibitem{ofner2017upper}
P.~Ofner, A.~Schwarz, J.~Pereira, and G.~R. M{\"u}ller-Putz, ``Upper limb
  movements can be decoded from the time-domain of low-frequency eeg,''
  \emph{PloS one}, vol.~12, no.~8, p. e0182578, 2017.

\bibitem{crone1998functional}
N.~E. Crone, D.~L. Miglioretti, B.~Gordon, and R.~P. Lesser, ``Functional
  mapping of human sensorimotor cortex with electrocorticographic spectral
  analysis. ii. event-related synchronization in the gamma band.''
  \emph{Brain}, vol. 121, no.~12, pp. 2301--2315, 1998.

\end{thebibliography}

% that's all folks
\end{document}